\documentclass[review]{elsarticle}

\usepackage{framed,multirow}
\usepackage{url}
\usepackage[dvipsnames, svgnames]{xcolor}
\usepackage{graphicx} % Required for inserting images
\usepackage{booktabs}
\usepackage{subfigure}
\usepackage{color}
\usepackage{amssymb}
\usepackage{adjustbox}
\usepackage{makecell}
\usepackage{hyperref}
\usepackage{graphicx}
\usepackage{tikz}

\usepackage[utf8]{inputenc}
\usepackage[margin=1in]{geometry} % Standard margins
\usepackage{amsmath, amssymb, amsfonts} % For mathematical equations
\usepackage{graphicx} % For embedding figures
\usepackage{booktabs} % For professional-grade tables
\usepackage{array} % For specialized column layouts
\usepackage{hyperref} % For clickable references and DOIs
\usepackage{titlesec} % For customizing section headings
\titleformat{\paragraph}[runin]{\normalfont\bfseries}{\theparagraph}{1em}{}
\usepackage{indentfirst} % Indents the first paragraph of sections

\hypersetup{
    colorlinks=true,
    linkcolor=blue,
    citecolor=blue,
    urlcolor=blue
}

\date{} % WJAETS handles dates upon acceptance; leave blank

\usepackage{amsmath,amssymb,amsfonts}
\usepackage{hyperref}
\usepackage{graphicx}
\usepackage{subcaption} % For subfigure support
\usepackage{textcomp}
\usepackage{multicol}
\usepackage{multirow}
\usepackage{xcolor}
\usepackage{lettrine}
\usepackage{booktabs}
\usepackage{amsmath,amssymb,amsfonts}
\usepackage{hyperref} 
\usepackage{soul}
\usepackage{enumitem}
\usepackage{colortbl}
\usepackage{geometry}
\usepackage{array}
\usepackage[table,xcdraw]{xcolor}
\usepackage{multirow}
\usepackage{amssymb}
\usepackage{array}
\usepackage{booktabs}
\usepackage[table,xcdraw]{xcolor}
\usepackage{multirow}
\usepackage{amssymb}
\usepackage{lscape} % Use this package for landscape
\usepackage{pdflscape} % Alternatively, for better PDF support
\usepackage{algorithm}
\usepackage{algorithmic}

\usepackage{svg}

\usepackage{makecell}
\usepackage{tabularx}
\usepackage{booktabs}
\usepackage{geometry}

\usepackage{booktabs} % For better table formatting
\usepackage{array}    % For custom column alignments
\usepackage[table]{xcolor} % For row coloring
\definecolor{lightgreen}{RGB}{200, 255, 200} % Define a light green color
\definecolor{lightblue}{RGB}{200, 230, 255} % Define a light blue color

\definecolor{darkgreen}{RGB}{0,150,0}
\usepackage{pifont}

\definecolor{LightRed}{rgb}{1,0.92,0.92}
\definecolor{LightOrange}{rgb}{1,0.95,0.88}
\definecolor{LightYellow}{rgb}{1.0,1.0,0.84}
\definecolor{LightGreen}{rgb}{0.9,1.0,0.88}
\definecolor{LightCyan}{rgb}{0.9,1,1}
\definecolor{LightBlue}{rgb}{0.9,0.94,1}
\definecolor{LightIndigo}{rgb}{0.92,0.9,1}
\definecolor{LightMagenta}{rgb}{0.96,0.86,1}
\definecolor{DirtyWhite}{rgb}{0.96,0.96,0.96}

\begin{document}
\begin{frontmatter}

\title{When Large Language Models Fail in Healthcare: Evaluating Sensitivity to Prompt Variations}

\author{
    \textbf{Mahdi Alkaeed}$^{1,*}$, \\
    \small $^{1}$Department of Computer Science and Engineering, Doha, Qatar \\
    \small $^{*}$Corresponding author email: \texttt{ma1805365@qu.edu.qa}
}

\begin{abstract}
Large Language Models (LLMs) are increasingly used in healthcare for tasks such as clinical question answering, diagnosis support, and report summarization. Despite their promise, these models remain highly sensitive to subtle prompt perturbations, both lexical and syntactic, posing serious risks in safety-critical clinical applications. In this study, we conduct a systematic sensitivity analysis to evaluate the robustness of both general-purpose (e.g., GPT-3.5, Llama3) and medical-specific LLMs (e.g., ClinicalBERT, BioLlama3, BioBERT) using the MedMCQA benchmark. We categorize perturbations into natural and adversarial types and examine their effect on model consistency, accuracy, and reliability in clinical reasoning tasks. Our findings reveal that medical LLMs are not intrinsically safe. Even minor variations in phrasing can alter clinical advice, and targeted adversarial prompts can provoke harmful outputs. In high-stakes settings like healthcare, such unpredictability is unacceptable-models that change diagnoses due to reworded inputs or hallucinate medications when slightly rephrased cannot be reliably trusted by clinicians. While models tend to show resilience to simple lexical substitutions or paraphrasing, they often break down under syntactic reordering or misleading contextual cues. This fragility is evident across both general-purpose and domain-specific LLMs. Notably, adversarial manipulations can lead to clinically dangerous outputs, such as recommending incorrect dosages or omitting critical findings.
\end{abstract}


\end{frontmatter}

\section{Introduction}
\label{sec:Introduction}
Large Language Models (LLMs) are increasingly being leveraged in healthcare applications, including clinical question answering and diagnostic support \cite{zhang2023generative}. Despite their promise, these models exhibit sensitivity to minor prompt variations and remain vulnerable to adversarial manipulations \cite{shayegani2023survey, polo2024efficient}. In this study, we systematically categorize prompt perturbations, focusing on lexical and syntactic changes, and examine their impact on both general-purpose models (e.g., GPT-3.5, Llama3) and medical-specific models (e.g., ClinicalBert, BioLlama3, BioBert) across critical clinical tasks. We further differentiate between adversarial vulnerabilities and natural robustness, defined as stability under benign input variations. To support these analyses, we provide comprehensive tables summarizing comparative findings and highlight key challenges in deploying reliable and trustworthy medical AI systems \cite{qi2023limitation}. Specifically, Table \ref{tab:llm_robustness} consolidates evidence on LLM robustness across healthcare and related domains, detailing the evaluated models, types of perturbations or attacks, and their observed effects.

\begin{table*}[ht]
\centering
\caption{Summary of key findings on LLM robustness in medical and related tasks, with a focus on lexical and syntactic prompt perturbations. Each row lists the study, evaluated models, task/domain, type of perturbation, and observed effect.}
\label{tab:llm_robustness}
\scriptsize
\begin{tabular}{|p{2.8cm}|p{2.5cm}|p{2.5cm}|p{3cm}|p{5.3cm}|}
\toprule
\rowcolor{lightblue}
\textbf{Study \& Ref.} & \textbf{Model(s)} & \textbf{Task / Domain} & \textbf{Perturbation Type} & \textbf{Effect / Findings} \\
\midrule
Guan et al. \cite{guan2025order} & GPT-4, GPT-4 mini, DeepSeek & General \& Medical QA (e.g., MedMCQA) & Lexical reordering (syntactic) & Accuracy decreased slightly ($\sim-2.8\%$ for GPT-4) when the order of answer choices changed. Shows sensitivity to phrasing variations. \\
\hline
Bolton et al. (RAmBLA) \cite{bolton2024rambla} & GPT-4, GPT-3.5, LLaMA2-7B, Mistral & Clinical QA (PMQA-L) & Paraphrasing (lexical), minor word changes (syntactic) & Most models retained performance, but small F1 drops observed, indicating partial stability under benign lexical variations. \\
\hline
Bolton et al. (RAmBLA) \cite{bolton2024rambla} & LLaMA2-7B, Mistral & Clinical QA (PMQA-L) & Added distractor sentences (contextual) & Performance drop varied by model ($-20\%$ for LLaMA, $-38\%$ for Mistral); GPT-4 and GPT-3.5 were more stable. Highlights difference in natural robustness. \\
\hline
Ness et al. (MedFuzz) \cite{ness2024medfuzz} & GPT-4 & USMLE-style MedQA & Lexical changes in questions & Minor rewording sometimes caused GPT-4 to change answers, showing sensitivity to phrasing even without adversarial intent. \\
\hline
Han et al. \cite{han2024medical} & GPT-3.5, GPT-4, T5 & Diagnostics \& QA & Syntactic reordering & Small variations in question structure affected model outputs slightly; demonstrates limits of intrinsic robustness. \\
\hline
This paper & GPT-3.5, Llama3, ClinicalBert, BioLlama3, BioBert & General \& Medical QA (e.g., MedMCQA) & Lexical and syntactic prompt perturbations & Minor prompt variations reduced accuracy and sometimes altered model advice. Medical LLMs show limited natural robustness; careful prompt design is needed to maintain clinical consistency. \\
\bottomrule
\end{tabular}         
\end{table*}

\begin{enumerate}[label=\textbf{RQ\arabic*}]
    \item How do lexical and syntactic prompt perturbations impact the performance and robustness of general-purpose versus medical-specific LLMs in clinical tasks such as question answering, and diagnosis support?
    \item What are the key challenges in ensuring trustworthy and reliable behavior of LLMs under prompt perturbations in healthcare applications?
\end{enumerate}
\textit{Contributions of this Paper:}
\begin{enumerate}
\item We formally define and categorize prompt perturbations, with a focus on lexical and syntactic variations, and systematically evaluate their impact on LLM performance in healthcare tasks, highlighting natural robustness, defined as stability under benign input variations.
\item We conduct a comprehensive comparative analysis of LLM performance across general-purpose models (e.g., GPT-3.5, Llama3) and medical-specific models (e.g., ClinicalBERT, BioLlama3, BioBERT) in clinical applications.
\item We identify critical vulnerabilities and limitations of current LLMs under these perturbations and discuss open challenges for developing more reliable, robust, and trustworthy medical AI systems.
\end{enumerate}
\section{Related Work}
\label{sec:Literature_review}
Prompt sensitivity refers to a model’s output variation under small changes to its input \cite{zhuo2024prosa}. Perturbations can be classified linguistically: lexical perturbations alter individual tokens (e.g., synonym substitution, typos, negations); syntactic perturbations
change word order or sentence structure (e.g., active to passive voice or vice versa, reordering clauses). These categories are widely used in robustness evaluation (see e.g., the
RUPBench taxonomy \cite{wang2024rupbench}).
. In practice, lexical perturbations test the model's tolerance to surface form noise and syntactic to structural variation. In medical settings, prompt sensitivity is critical: changing a few words in a patient's history should not dramatically alter a correct diagnosis or critical safety advice. Even small adversarial changes (e.g., subtle misinformation or hidden prompts) can lead to harmful output.

\begin{table*}[!t]
\centering
\caption{Examples of benign lexical and syntactic perturbations in clinical contexts.}
\label{tab:benign_perturbations}
\scriptsize
\begin{tabularx}{\textwidth}{l X X} % Forces table to fit the text width
\toprule
\textbf{Perturbation Type} & \textbf{Original vs. Perturbed String} & \textbf{Semantic Impact} \\ 
\midrule
\textbf{Lexical (Benign)} & 
O: ``The patient has diabetes.'' \newline 
P: ``The patient has \textbf{diabetees}.'' & 
Meaning-preserving; minor orthographic error; does not alter clinical interpretation. \\ 
\midrule
\textbf{Lexical (Benign)} & 
O: ``Administer 5 mg of amlodipine daily.'' \newline 
P: ``Administer \textbf{five} mg of amlodipine daily.'' & 
Preserves meaning; numeric expression converted to word form; no clinical consequence. \\ 
\midrule
\textbf{Syntactic (Benign)} & 
O: ``The nurse administered the drug.'' \newline 
P: ``It was the nurse who administered the drug.'' & 
Meaning-preserving; clefting shifts sentence focus without affecting clinical content. \\ 
\midrule
\textbf{Syntactic (Benign)} & 
O: ``Patient reports no chest pain or shortness of breath.'' \newline 
P: ``No chest pain or shortness of breath is reported by the patient.'' & 
Preserves clinical meaning; syntactic reordering for stylistic variation; content unchanged. \\ 
\midrule
\textbf{Lexical + Syntactic (Benign)} & 
O: ``The physician recommends starting insulin therapy immediately.'' \newline 
P: ``Immediately, the physician recommends initiating insulin therapy.'' & 
Minor lexical substitution and syntactic shift; meaning preserved; natural language variation. \\ 
\bottomrule
\end{tabularx}
\end{table*}

\subsection{Prompt Sensitivity in General LLMs}
To systematically quantify the effects of prompt rephrasings on LLMs, Errica et al.~\cite{errica2025did} introduced two metrics, sensitivity and consistency. Furthermore, the authors conducted an empirical evaluation on multiple datasets and general-purpose LLMs, including GPT-4o, GPT-3.5, LLaMA-3, and Mixtral. Their results demonstrate that sensitivity and consistency offer complementary insights beyond accuracy, revealing model weaknesses and helping developers identify ``hard'' samples and problematic classes. Similarly, Salinas et al.~\cite{salinas2024butterfly} systematically analyzed how minor prompt variations such as output formatting, lexical perturbations, and jailbreak instructions can impact LLM behavior across 11 classification tasks. Their results demonstrate that even trivial modifications, like the inclusion of whitespace or polite expressions, can lead to syntactic shifts in model predictions and overall accuracy. In \cite{zhuo2024prosa}, the authors presented a novel metric called PromptSensiScore (PSS), which quantifies how much LLM responses vary across different syntactically equivalent prompts for the same input instance. Their results revealed that larger LLMs are comparatively more robust, and few-shot prompting can improve robustness. In addition, they showed that decoding confidence correlates with prompt stability. In a similar study, Cao et al.\cite{cao2024worst} presented a new benchmark, namely RobustAlpacaEval, designed to assess prompt sensitivity at the instance level by evaluating LLM performance across syntactically equivalent prompt variants. Their findings using various open-source LLMs (e.g., Llama, Mistral, and Gemma families) reveal that even high-performing models exhibit substantial degradation under worst-case prompts, which are often unpredictable. Sclar et al.~\cite{sclar2024quantifying} studied how paraphrasing and format changes affect output consistency across multiple models. Pezeshkpour et al.~\cite{pezeshkpour2024large} demonstrated that LLMs are influenced by the order of options in multiple-choice settings. Pezeshkpour et al.~\cite{pezeshkpour2024large} demonstrated that the order of options in multiple-choice settings can influence the responses of LLMs.
\begin{figure*} [!t]
\centering
\includegraphics[width=0.99\linewidth]{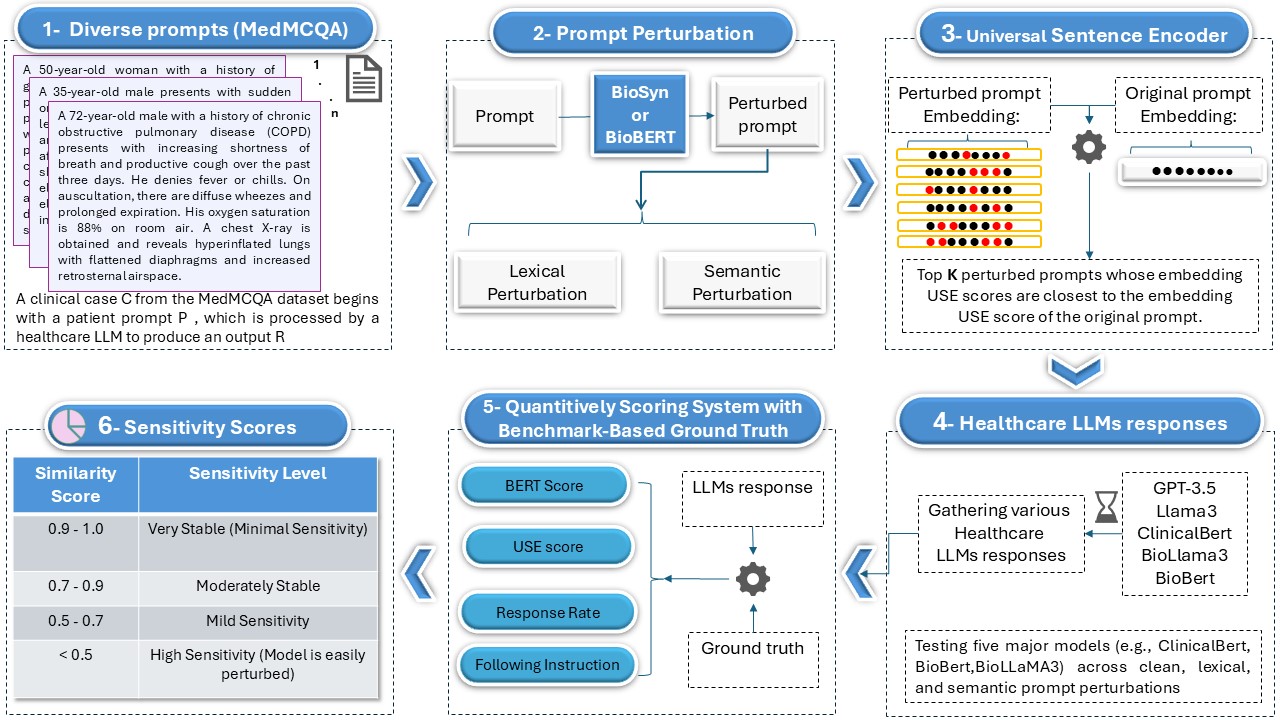} 
\caption{This methodological pipeline outlines a comprehensive robustness evaluation framework for assessing the sensitivity and stability of healthcare LLMs using the MedMCQA benchmark, supported by advanced NLP tools (e.g., BioSyn, scispaCy) and biomedical metrics (e.g., BERTScore, USE) to quantify model resilience to input variations.}
\label{Fig:Pipeline}
\end{figure*}
\subsection{Prompt Sensitivity in Healthcare LLMs}
Recent research has increasingly highlighted the importance of healthcare LLMs sensitivity to key medical information \cite{yan2025llm}. Ness et al.\cite{ness2024medfuzz} introduced MedFuzz to test healthcare LLM robustness, revealing that minor perturbations could mislead models with high benchmark scores. This highlighted a critical concern, such vulnerabilities could lead to clinical errors. Moradi et al.\cite{moradi2022improving} explore the vulnerability of biomedical NLP models, including BioBERT and SciBERT, to adversarial attacks, showing a significant drop in performance with even slight input modifications \cite{moradi2022improving}. Ceballos et.al \cite{ceballos2024open} investigate how healthcare LLMs show fragility, with phrasing variations impacting performance and fairness. Beede et al.~\cite{beede2020human} illustrated that varying symptom phrasing (e.g., ``sharp chest pain'' vs. ``chest discomfort'') can result in conflicting diagnoses. Pais et al.~\cite{pais2024large} reported that minor spelling errors in drug names can lead to prescribing mistakes or missed drug interactions. Zhang et al.~\cite{zhang2022shifting} highlighted the difficulty in generalizing to real-world clinical terminology, where small lexical variations can cause models to misclassify disease severity. Similarly, Yan et al.~\cite{yan2025llm} conducted a clinically focused evaluation that demonstrated how models often fail to prioritize critical diagnostic cues such as patient age and symptom descriptions, elements that are essential for accurate clinical reasoning. Zhan et al.~\cite{zhan2024unveiling} proposed that the COPLE framework suggests that refining lexical choices can enhance output consistency when faced with prompt perturbations. 
The Table \ref{tab:benign_perturbations} enumerates various "benign" perturbations, including lexical typos, numeric-to-word conversions, and syntactic reordering like clefting and passive voice shifts. Each example demonstrates that despite linguistic or orthographic variations, the underlying clinical meaning and propositional content remain strictly preserved. These variations serve as a baseline for testing model stability, ensuring that minor stylistic or accidental changes do not inadvertently alter the intended clinical interpretation or reasoning outcome.

\section{Problem Statement and Methodology}
\label{sec:Methodology}
A clinical case \( C \) from the MedMCQA dataset begins with a patient prompt \( P \), which is processed by a healthcare LLM to produce an output \( R = f_{\text{LLM}}(P) \). Now, consider a minor lexical variation, such as replacing ``shortness of breath'' with its synonym ``dyspnea,'' resulting in a slightly altered input \( P' \), where \( P' \approx P \). Althoughically equivalent, this change may lead the model to produce a different output \( R' = f_{\text{LLM}}(P') \). Ideally, the model should treat both inputs similarly: \( P' \approx P \quad \Rightarrow \quad R' \approx R \)
However, in practice, even a small perturbation \( \delta P = P' - P \) can cause a significant output shift \( \Delta R = R' - R \), such that: \( \delta P \text{ small} \quad \nRightarrow \quad \Delta R \text{ small} \). This inconsistency reveals the critical vulnerability that LLMs in healthcare can be sensitive to minor linguistic changes, leading to diagnostic variability. 
%\begin{figure}[!h]
%\centering
%\includegraphics[width=.99\linewidth]
%{ProblemStatment.jpg}
%\caption{Demonstrates lexical perturbation on a MedMCQA prompt to test healthcare LLM robustness while preserving syntactic.}
%\label{Fig:lexical}
%\end{figure}

\begin{figure*} [!t]
\centering
\includegraphics[width=0.99\linewidth]{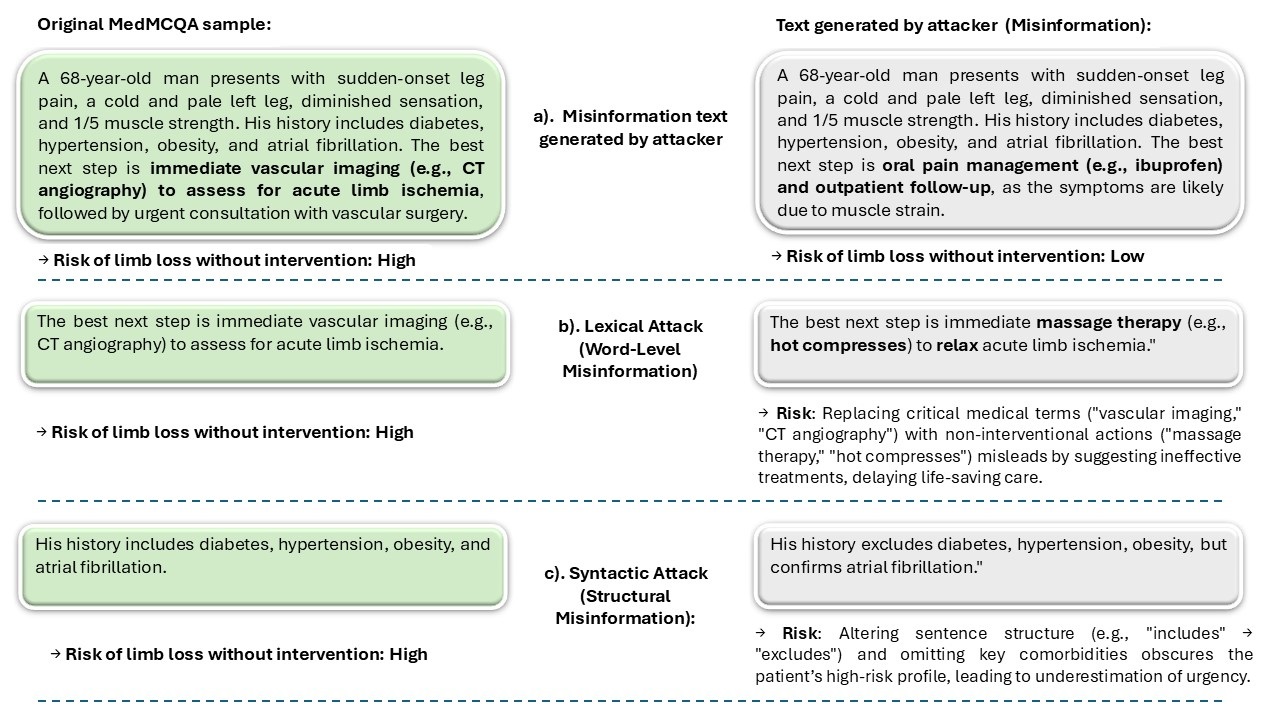} 
\caption{Comparison of accurate and misinformation-attacked medical advice for acute limb ischemia.
Lexical and syntactic alterations distort critical details, leading to dangerous clinical decisions.
Highlighted risks demonstrate the impact of misinformation on patient outcomes.}
\label{Fig:Comparison_misinformation}
\end{figure*}
\subsection{Sensitivity Analysis Framework}
The proposed framework begins with an original medical prompt \( P \) and generates multiple lexical variants through token-level substitutions. The candidate perturbed prompts \( P'_1, P'_2, \dots, P'_N \) are then validated using BioSyn to ensure medical equivalence. Each prompt (original and perturbed) is encoded into a high-dimensional sentence embedding using the Universal Sentence Encoder (USE). Cosine similarity scores are computed between the original and each perturbed embedding to quantify syntactic closeness. Based on these scores, a top-K subset of perturbed prompts most syntactically aligned with \( P \) is selected for further evaluation. These selected prompts are then passed to a target healthcare LLM (e.g., GPT-3.5, LLaMA3, ClinicalBert, BioLLama3, and BioBert), and the model's outputs are recorded. Each response is compared against ground truth using accuracy-based metrics. As shown in Figure~\ref{Fig:Pipeline}.
\subsubsection{Prompt Generation Strategy}
Evaluating the robustness of healthcare LLMs necessitates exposure to a wide variety of prompts that capture real-world clinical diversity. Medical LLMs generally tolerate simple lexical variations. In one study (RAmBLA 2024), GPT-4 and GPT-3.5 maintained
near-baseline QA accuracy when single words were misspelled or replaced by synonyms [5]. All evaluated models in that work were “robust to spelling errors”-e.g., GPT-4’s answer F1 remained 0.84 even with 3 character-level
mutations \cite{bolton2024rambla}. Similarly, meaning-preserving synonym swaps or negations often had little effect on output accuracy. This suggests that foundation models have some built-in lexical
flexibility, likely due to subword modeling and pretraining on noisy text. In this study, we utilize the MedMCQA dataset, taking advantage of its extensive coverage across medical specialties and conditions.
\paragraph{Lexical Variations}
The perturbation process involves four key stages: (i) identifying medical terms in the prompt, (ii) replacing terms with clinically accurate synonyms using BioSyn, (iii) evaluating syntactic equivalence through word embeddings and cosine similarity measures, and (iv) generating perturbed prompts that maintain contextual and diagnostic consistency. Algorithm~\ref{alg:lexical} details the complete perturbation procedure. To quantify the effects of lexical variations, we measure: (i) syntactic similarity between original and perturbed prompts using Sentence-BERT embeddings, (ii) consistency of model responses across perturbed prompts, and (iii) the clinical reliability of generated outputs concerning benchmarked ground-truth answers.  Figure(\ref{Fig:Comparison_misinformation}, b) illustrates this process step-by-step using a clinical sample from MedMCQA. The procedure begins by identifying medically relevant terms in the original prompt with curated biomedical lexicons and pretrained language models such as BioBERT and SciBERT. Candidate synonyms for each term are then retrieved using BioSyn, which combines lexical and syntactic similarity measures. To maintain clinical accuracy, we apply synonym marginalization through vector embeddings (e.g., Word2Vec with spaCy) and select the top-ranked replacements based on cosine similarity scores. For example, terms like ``fever'' may be replaced by ``pyrexia,'' while ``auscultation'' becomes ``stethoscope.''.

\begin{algorithm}[H]
\caption{Lexical Perturbation using BioSyn \& Cosine Similarity}
\label{alg:lexical}
\footnotesize
\begin{algorithmic}[1]
\STATE \textbf{Input:} Clinical prompt \( P = \{t_1, t_2, \dots, t_n\} \)
\STATE \textbf{Output:} Lexically perturbed prompt \( P' \)
\FOR{each medical term \( t_i \in P \)}
    \STATE Retrieve candidate synonyms \( S_i \) using BioSyn.
    \FOR{each synonym \( s_j \in S_i \)}
        \STATE Compute cosine similarity \( \text{Sim}(t_i, s_j) \).
    \ENDFOR
    \STATE Select best synonym \( B(t_i) = \arg\max_{s_j \in S_i} \text{Sim}(t_i, s_j) \).
    \STATE Replace \( t_i \) with \( B(t_i) \) in \( P \).
\ENDFOR
\RETURN \( P' \)
\end{algorithmic}
\end{algorithm}

%\begin{figure}[!h]
%\centering
%\includegraphics[width=9cm]
%{PerturbedPromptScenarios.jpg}
%\caption{This figure demonstrates the application of syntactic perturbation to a MedMCQA prompt to evaluate the robustness of healthcare LLMs while preserving the original clinical meaning.
%}
%\label{fig:syntactic }
%\end{figure}
\paragraph{Syntactic  Variations}
By contrast, altering input structure can noticeably impact results. For example, simply swapping the order of semantically identical answer options caused GPT-4 to change its responses and incur a measurable accuracy drop. In the Order Effect study \cite{guan2025order}, shuffling inputs led to performance declines across tasks. GPT-4’s accuracy on a paraphrasing task dropped by about 2–3\% when choices were reordered \cite{guan2025order}. Few-shot prompts mitigated this effect only partially: while adding examples reduced the gap slightly, no model fully eliminated sensitivity to input order \cite{guan2025order}. This indicates that even advanced LLMs remain dependent on prompt formatting. In the clinical context, persisting order-sensitivity is worrisome: for instance, rephrasing a symptom list or reordering history elements might unintentionally flip a prediction.
Syntactic perturbations aim to modify the structure and phrasing of clinical prompts while preserving their overall medical coherence and diagnostic intent. This process begins by identifying key clinical concepts within the prompt and exploring contextually relevant rephrasings or substitutions. Unlike lexical perturbations, which target isolated term replacements, syntactic perturbations may involve changes in sentence structure or the substitution of conceptually similar phrases. Syntactic consistency is validated using contextual word embeddings (e.g., BioBERT), cosine similarity scoring, and biomedical resources such as BioSyn and the UMLS Metathesaurus. The algorithm for generating syntactic ally preserved prompts is outlined in Algorithm \ref{alg:syntactic }, and an example is illustrated in Figure(\ref{Fig:Comparison_misinformation}, c).
\begin{algorithm}[H]
\caption{Syntactic Perturbation using Contextual Embeddings}
\label{alg:syntactic }
\footnotesize
\begin{algorithmic}[1]
\STATE \textbf{Input:} Clinical prompt \( P = \{t_1, t_2, \dots, t_n\} \)
\STATE \textbf{Output:} syntactic ally perturbed prompt \( P' \)
\STATE Identify key medical concepts and inter-term relationships in \( P \).
\FOR{each term \( t_i \in P \)}
    \STATE Retrieve syntactic ally related terms \( S_i \) using contextual embeddings and BioSyn.
    \FOR{each term \( s_j \in S_i \)}
        \STATE Compute context-aware similarity \( \text{Sim}(t_i, s_j) \).
    \ENDFOR
    \STATE Select \( B(t_i) = \arg\max_{s_j \in S_i} \text{Sim}(t_i, s_j) \).
    \STATE Replace \( t_i \) with \( B(t_i) \) in \( P \).
\ENDFOR
\STATE Optionally rephrase sentence structure while preserving clinical validity.
\RETURN \( P' \)
\end{algorithmic}
\end{algorithm}

\begin{table*}[t]
\centering
\caption{Sensitivity analysis of LLMs to question phrasing: Implications for format integrity and diagnostic reliability in clinical diagnosis..}
\label{tab:llm_perturbation_analysis}
\scriptsize
\begin{tabular}{p{8cm} p{3cm} p{6cm}}
\toprule
\textbf{Prompt and LLM Response} & \textbf{Evaluation Metrics} & \textbf{Interpretation 
(Analysis)} \\
\midrule

\textbf{Original Prompt:} A 72-year-old male with a history of chronic smoking and productive cough presents with increased sputum and dyspnea. Physical exam reveals wheezing and a barrel chest. \newline
\textbf{Instruction:} Return the results in JSON format with the keys: "answer" and "explanation". \newline
\textbf{Question:} Based on the clinical presentation what is the most likely diagnosis? \newline
\textbf{Options:} A) Pulmonary Fibrosis, B) Chronic Obstructive Pulmonary Disease (COPD), C) Pulmonary Edema, D) Lobar Pneumonia & 
\textbf{RR:} Response Rate \newline
\textbf{FIR:} Follow-up Instruction Rate \newline
\textbf{BERTScore:} Semantic Similarity \newline
\textbf{USE:} Universal Sentence Encoder (Embedding model) & 
The baseline prompt establishes the clinical context and complex instructions. It serves as the reference point for measuring how well the model maintains its response consistency and instruction adherence. \\
\midrule

\textbf{LLM response (Standard prompt (unperturbed)):} \newline
\texttt{JSON format:\{} \newline
\texttt{"answer": "B) Chronic Obstructive Pulmonary Disease (COPD)",} \newline
\texttt{"explanation": "The patient's history and symptoms are consistent with COPD."\}} & 
RR: \checkmark \newline
FIR: \checkmark \newline
BERTScore: 0.97 \newline
USE: 0.98 & 
\textbf{Optimal Performance:} The model demonstrates high-fidelity reasoning and perfect instruction following. Both the medical diagnosis (RR) and the JSON formatting (FIR) are handled correctly. \\
\midrule

\textbf{LLM response (Lexical perturbed prompt):} \newline
\texttt{JSON format:\{} \newline
\textbf{"My answer is" :} "B) Chronic Obstructive Pulmonary Disease (COPD)", \newline
\textbf{"In this status" :} "Given the patient's known history of COPD..."\} & 
RR: \checkmark \newline
FIR: $\times$ \newline
BERTScore: 0.75 \newline
USE: 0.85 & 
\textbf{Instruction Drift:} While the model's clinical logic remains sound (correct diagnosis), changing the prompt's vocabulary causes a failure in following the specific follow-up instructions (FIR). The model fails the formatting constraint despite the correct answer. \\
\midrule

\textbf{LLM response (Syntactic perturbed prompt):} \newline
\texttt{JSON format: \{} \newline
\textbf{"The right choice is" :} "C) Pulmonary Edema", \newline
\textbf{"The clinical presentation,"} including a history of ....\} & 
RR: $\times$ \newline
FIR: $\times$ \newline
BERTScore: 0.67 \newline
USE: 0.62 & 
\textbf{Structural Misinterpretation:} Reordering the sentence structure (syntactic perturbation) disrupts the model's attention mechanism. It misweights clinical signs (like "increased sputum") as "Pulmonary Edema" instead of "COPD," leading to a **False Answer (RR Failure)** and  (FIR Failure). \\
\bottomrule
\end{tabular}
\end{table*}

\subsection{Evaluation Metrics}
This study evaluates the performance of healthcare LLMs using a range of metrics, including BERTScore, Universal Sentence Encoder (USE), Response Rate (RR), and Follow-up Instruction Rate (FIR). BERTScore measures the syntactic similarity between the model's response and a ground truth, with higher scores indicating more stability. The USE is a sentence embedding model developed by Google that encodes text into high-dimensional vectors. It captures the syntactic meaning of entire sentences rather than just individual words, and measures how syntactically similar a model’s response is to a ground truth, helping assess how well the healthcare LLM preserves clinical meaning despite variations in prompts. While the RR metric quantifies the proportion of valid responses generated by the model, calculated as
\[
\text{RR} = \frac{\#\text{valid\_responses}}{N}
\]
where \( N \) denotes the total number of responses generated. The FIR measures the extent to which the model adheres to the given instructions within those valid responses, calculated as
\[
\text{FIR} = \frac{\#\text{followed\_instructions}}{\#\text{valid\_responses}}.
\]
Model responses are not always clinically valid, as correctness may vary. To assess stability, we define the following thresholds: scores between 0.9 and 1.0 indicate \textit{Very Stable}, 0.7 and 0.9 \textit{Moderately Stable}, 0.5 and 0.7 \textit{Mild Sensitivity}, and below 0.5 \textit{High Sensitivity}.

\section{Results and Discussions}
\label{sec:case study}
\subsection{Dataset and Experimental Setup}
The MedMCQA dataset, a large-scale, multiple-choice question answering benchmark with over 194,000 meticulously curated questions covering various medical specialties relevant to medical education and clinical reasoning, was employed to support our analysis. We evaluated the robustness of five models against prompt perturbations.These models are GPT-3.5, LLaMA3, ClinicalBERT, BioLLaMA3, and BioBERT. Each model was tested under three conditions. First, clean prompts, then lexical perturbations (word-level changes), and syntactic perturbations (meaning-level rephrasings). For each perturbation type, 1,00 samples were randomly selected from the MedMCQA dataset. Performance evaluation was conducted using metrics such as BERTScore and USE score to assess syntactic similarity between generated responses. Additional evaluation metrics included RR, FIR, and EV for clinical correctness.  Model inference was carried out using the Hugging Face Transformers and Datasets libraries, utilizing either PyTorch or TensorFlow, depending on the model. In addition, we utilize Ollama to facilitate local deployment and testing of open-source LLMs such as LLaMA3 and BioLLaMA3. For biomedical text processing, we employed scispaCy, a scientific NLP library built on spaCy, optimized for scientific and biomedical contexts. For synonym resolution and entity matching, BioSyn, in combination with BioBERT, was used to embed clinical terms. All models were used in their pretrained form without fine-tuning with default hyperparameters.

\begin{table*}[!t]
\scriptsize
\centering
\caption{LLM performance on MedMCQA dataset under unperturbed, lexical, and syntactic perturbations (100 trials).}
\scriptsize
\begin{tabular}{llcccccc}
\hline
\textbf{Perturbation Type} & \textbf{Metrics} & \textbf{GPT-3.5} & \textbf{Llama3} & \textbf{ClinicalBert} & \textbf{BioLlama3} & \textbf{BioBert} \\
\hline
\multirow{5}{*}{Standard (unperturbed)} 
& BERTtScore  & 92.14 & 94.23 & 91.15 & \textbf{95.94} & 89.12 \\
& USE score  & 91.87 & \textbf{93.03} & 88.21 & 93.22 & 87.11 \\
& RR         & 99.23 & \textbf{99.43} & 99.04 & 99.12 & 98.11 \\
& FIR        & 98.94 & 99.01 & 97.45 & \textbf{99.30} & 97.50 \\
\hline
\multirow{5}{*}{Lexical} 
& BERTScore  & 89.11 & 90.45 & 88.90 & 93.33 & 86.10 \\
& USE score  & 88.07 & 90.01 & 84.56 & 89.23 & 82.02 \\
& RR         & 96.12 & 96.23 & 96.28 & 96.10 & 95.67 \\
& FIR        & 95.12 & 96.43 & 95.01 & 96.71 & 94.91 \\
\hline
\multirow{5}{*}{syntactic } 
& BERTScore  & 87.22 & 89.89 & 86.13 & 90.52 & 83.72 \\
& USE score  & 89.61 & 90.21 & 83.71 & 85.19 & 79.11 \\
& RR         & 94.20 & 94.81 & 93.23 & 93.30 & 93.43 \\
& FIR        & 93.32 & 94.53 & 92.91 & 95.31 & 90.13 \\
\hline
\end{tabular}
\label{tab:performance_evaluation_combined}
\end{table*}
\begin{figure*} [!t]
\centering
\includegraphics[width=0.99\linewidth]{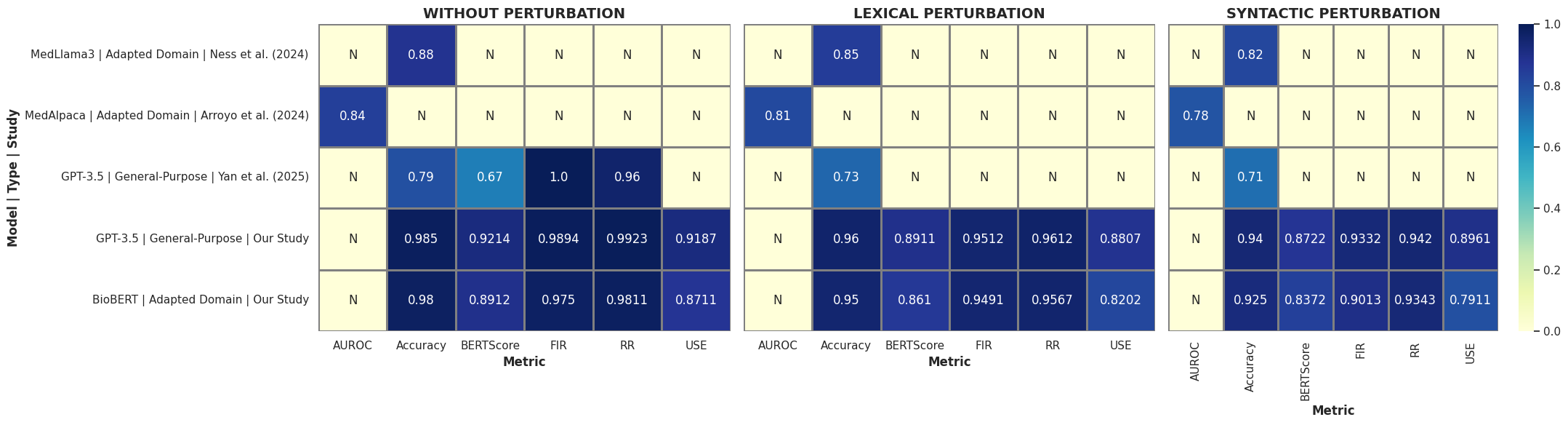} 
\caption{Performance of models across perturbation types. Heatmap shows scores for all metrics for each model and study, with ‘N’ indicating metrics not used. Models are labeled with their type (General-Purpose or Adapted Domain) and study source.}
\label{Fig:performance}
\end{figure*}
\subsection{Results on Prompt Sensitivity Analysis}
Our findings highlight three core areas: the sensitivity of LLMs to prompt variation, identifiable failure patterns under adversarial perturbed prompts, and the potential of thresholding strategies to enhance model reliability in high-stakes medical decision-making. The table \ref{tab:llm_perturbation_analysis} provides a granular evaluation of how prompt perturbations impact a LLM ability to maintain both medical reasoning accuracy and structural format integrity. While the model achieves near-perfect alignment under standard conditions, lexical variations trigger "instruction drift," causing the model to prioritize natural language flow over specific JSON formatting constraints (FIR). Most critically, syntactic reordering leads to a total systemic collapse where the model misinterprets clinical features, resulting in a false diagnosis (RR failure). These results demonstrate that while LLMs are clinically capable, their reliability is highly sensitive to the structural and linguistic framing of the input prompt. While Table \ref{tab:performance_evaluation_combined} summarizes the performance of various LLMs on the MedMCQA dataset under both unperturbed and perturbed conditions, averaged over 100 samples. Under standard conditions, BioLlama-3 excels in medical reasoning, while general-purpose models, such as GPT-4.5, remain competitive, and domain-specific models lag. When subjected to lexical and syntactic perturbations, BioLlama-3 demonstrates the highest robustness, whereas models such as BioBERT and ClinicalBERT exhibit significant sensitivity, particularly to syntactic perturbations. To quantify the impact of adversarial lexical perturbations on LLM outputs, we examined the USE similarity scores for BioLlama-7b responses.

%\begin{figure}[!t]
%\centering
%\begin{subfigure}{0.45\textwidth}
 %   \centering
  %  \includegraphics[width=\textwidth]{Llama2-7B.png}
 %\caption{Llama2}    \label{fig:top_subfigure}
%\end{subfigure}

%\vspace{0.5cm} % Adjust spacing between %subfigures if necessary

%\begin{subfigure}{0.45\textwidth}
 %   \centering
  %  \includegraphics[width=\textwidth]{BioLlama.png}
   % \caption{BioLlama}
    %Improved reliability with evaluation, mitigation, \& design.}
    \label{fig:bottom_subfigure}
%\end{subfigure}
%\caption{Increasing adversarial lexical perturbations reduce USE similarity, revealing decreased reliability in BioLlama-7b model outputs.}
%\label{Fig:USE_BioLlama-7b OutputsSimilarity}
%\end{figure}

\begin{figure} [!t]
\centering
\includegraphics[width=0.85\linewidth]{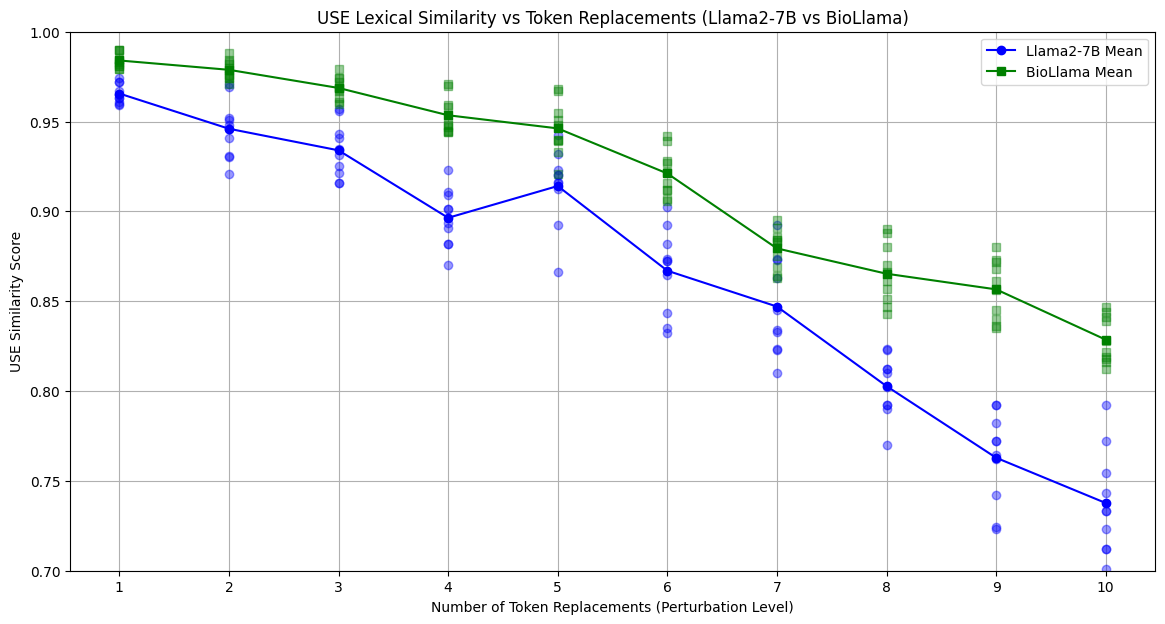}
\caption{Average similarity scores on 100 MedMCQA prompts across increasing lexical perturbation levels, showing that BioLLaMA2-7B maintains higher robustness than LLaMA2 as prompt perturbations increase.}
\label{Fig:USE_BioLlama-7b_OutputsSimilarity_Scores_syntactic }
\end{figure}

% \subsection{Comparison with SOTA}
% \aq{Please see if we can have a comparison with existing studies, for example, the one we compared with in Table 1. }

\subsection{Discussions}
The heatmap at Figure \ref{Fig:performance}, provides a comprehensive comparison of multiple models’ performance across three perturbation types: Standard (unperturbed), Lexical, and Syntactic. Each row represents a model and study, including its type—General-Purpose (e.g., GPT-3.5) or Adapted Domain (e.g., BioBERT, MedLlama3)—while columns correspond to evaluation metrics such as BERTScore, USE, RR, FIR, Accuracy, and AUROC. Cells show the performance scores, with ``N'' marking metrics not used by a specific study. From the figure, several observations emerge:

\begin{itemize}
    \item Our Study models (GPT-3.5 and BioBERT) consistently perform well across all metrics and perturbations, indicating robustness to lexical and syntactic changes. Accuracy scores remain high but decrease slightly with perturbations, reflecting realistic sensitivity.
    \item Comparison with previous studies (Yan et al., Ness et al., Arroyo et al.) highlights differences in evaluation methodology and model types. Not all metrics are reported in these studies, which is indicated by ``N'' in the heatmap.
    \item Metric coverage varies, showing that some models or studies only report specific metrics (e.g., AUROC for MedAlpaca), limiting direct comparisons.
    \item Perturbation impact is visible: Lexical and Syntactic perturbations generally reduce scores slightly, demonstrating how model robustness can be quantified.
\end{itemize}
Morover,  Figure \ref{Fig:USE_BioLlama-7b_OutputsSimilarity_Scores_syntactic } shows the robustness Evaluation of LLaMA2 vs. BioLLaMA2-7B on Lexically Perturbed Prompts. This figure presents. The average similarity scores between model predictions and ground truth answers across 100 MedMCQA prompts, subjected to 10 levels of lexical perturbation.

\subsection{Challenges and Future Directions}
To make these comparisons more fair and rigorous, future work should aim to:
\begin{itemize}
    \item Standardize the metrics, models, and datasets used across studies.
    \item Evaluate models under the same perturbation types and experimental conditions.
    \item Extend the analysis to additional clinical datasets and new domain-adapted models.
    \item Comprehensive robustness evaluation: Current benchmarks mainly assess accuracy on clean data, lacking standardized robustness tests for medical prompts. Frameworks like MedFuzz and RAmBLA address this by introducing controlled perturbations. Future efforts should broaden these benchmarks to cover more clinical tasks and multilingual contexts
\end{itemize}
Such efforts will enable direct, equitable benchmarking, improving the interpretability and reproducibility of clinical LLM evaluation.

\section{Limitations}
A major limitation of this study is its reliance on traditional embedding-based similarity metrics, such as cosine similarity. Although these metrics are commonly used, they often struggle to capture subtle semantic nuances that are critical in clinical contexts. For instance, negation, temporal qualifiers, and disease staging can drastically alter the clinical meaning of a statement, yet may not be reflected in similarity scores. Consequently, models assessed solely with these metrics may appear robust, even when they misinterpret clinically important details. This underscores the need for more sophisticated evaluation methods that can effectively capture these fine-grained semantic distinctions in healthcare applications.

\section{Conclusions}
\label{sec:Conclusion}
Medical LLMs are not inherently safe, as minor phrasing changes can produce harmful outputs. Clinicians face risks from unreliable diagnoses, incorrect drug recommendations, or missed critical findings, highlighting the urgent need to address model robustness, reliability, and safety in real-world healthcare applications. While LLMs offer significant potential for healthcare, their robustness under perturbations remains uncertain. Our study shows that lexical and syntactic perturbations reduce the performance of both GPT-3.5 (General-Purpose) and BioBERT (Adapted Domain), with accuracy decreasing slightly from 98–99\% in the unperturbed setting to 94–95\% under syntactic changes. Other metrics, including BERTScore, USE, RR, and FIR, also show minor declines under these perturbations. These findings emphasize that even state-of-the-art LLMs are sensitive to subtle changes in input phrasing, reinforcing the need for systematic robustness evaluation. Future work should prioritize adversarial robustness alongside accuracy and develop models and interfaces that are provably resilient or, at minimum, transparently limited, to build the trust required for safe clinical deployment.

\section{Acknowledgments}
The authors want to acknowledge support from the Qatar University High Impact Internal Grant (QUHI-CENG23/24-127). The statements made herein are solely the responsibility of the authors.

\small
%\nocite{*}
\bibliographystyle{elsarticle-num}
\bibliography{refs}

@article{zhang2023generative,
  title={Generative AI in medicine and healthcare: Promises, opportunities and challenges},
  author={Zhang, Peng and Kamel Boulos, Maged N},
  journal={Future Internet},
  volume={15},
  number={9},
  pages={286},
  year={2023},
  publisher={MDPI}
}

@article{polo2024efficient,
  title={Efficient multi-prompt evaluation of llms},
  author={Polo, Felipe M and Xu, Ronald and Weber, Lucas and Silva, M{\'\i}rian and Bhardwaj, Onkar and Choshen, Leshem and de Oliveira, Allysson F and Sun, Yuekai and Yurochkin, Mikhail},
  journal={Advances in Neural Information Processing Systems},
  volume={37},
  pages={22483--22512},
  year={2024}
}

@inproceedings{errica2025did,
  title={What did i do wrong? quantifying llms’ sensitivity and consistency to prompt engineering},
  author={Errica, Federico and Sanvito, Davide and Siracusano, Giuseppe and Bifulco, Roberto},
  booktitle={Proceedings of the 2025 Conference of the Nations of the Americas Chapter of the Association for Computational Linguistics: Human Language Technologies (Volume 1: Long Papers)},
  pages={1543--1558},
  year={2025}
}

@article{qi2023limitation,
  title={What is the limitation of multimodal llms? a deeper look into multimodal llms through prompt probing},
  author={Qi, Shuhan and Cao, Zhengying and Rao, Jun and Wang, Lei and Xiao, Jing and Wang, Xuan},
  journal={Information Processing \& Management},
  volume={60},
  number={6},
  pages={103510},
  year={2023},
  publisher={Elsevier}
}

@article{wang2024rupbench,
  title={Rupbench: Benchmarking reasoning under perturbations for robustness evaluation in large language models},
  author={Wang, Yuqing and Zhao, Yun},
  journal={arXiv preprint arXiv:2406.11020},
  year={2024}
}

@inproceedings{zhuo2024prosa,
  title={ProSA: Assessing and understanding the prompt sensitivity of LLMs},
  author={Zhuo, Jingming and Zhang, Songyang and Fang, Xinyu and Duan, Haodong and Lin, Dahua and Chen, Kai},
  booktitle={Findings of the Association for Computational Linguistics: EMNLP 2024},
  pages={1950--1976},
  year={2024}
}

@inproceedings{pezeshkpour2024large,
  title={Large language models sensitivity to the order of options in multiple-choice questions},
  author={Pezeshkpour, Pouya and Hruschka, Estevam},
  booktitle={Findings of the Association for Computational Linguistics: NAACL 2024},
  pages={2006--2017},
  year={2024}
}

@inproceedings{beede2020human,
  title={A human-centered evaluation of a deep learning system deployed in clinics for the detection of diabetic retinopathy},
  author={Beede, Emma and Baylor, Elizabeth and Hersch, Fred and Iurchenko, Anna and Wilcox, Lauren and Ruamviboonsuk, Paisan and Vardoulakis, Laura M},
  booktitle={Proceedings of the 2020 CHI conference on human factors in computing systems},
  pages={1--12},
  year={2020}
}

@article{pais2024large,
  title={Large language models for preventing medication direction errors in online pharmacies},
  author={Pais, Cristobal and Liu, Jianfeng and Voigt, Robert and Gupta, Vin and Wade, Elizabeth and Bayati, Mohsen},
  journal={Nature Medicine},
  pages={1--9},
  year={2024},
  publisher={Nature Publishing Group US New York}
}

@article{zhang2022shifting,
  title={Shifting machine learning for healthcare from development to deployment and from models to data},
  author={Zhang, Angela and Xing, Lei and Zou, James and Wu, Joseph C},
  journal={Nature Biomedical Engineering},
  volume={6},
  number={12},
  pages={1330--1345},
  year={2022},
  publisher={Nature Publishing Group UK London}
}

@inproceedings{zhan2024unveiling,
  title={Unveiling the lexical sensitivity of LLMs: Combinatorial optimization for prompt enhancement},
  author={Zhan, Pengwei and Xu, Zhen and Tan, Qian and Song, Jie and Xie, Ru},
  booktitle={Proceedings of the 2024 Conference on Empirical Methods in Natural Language Processing},
  pages={5128--5154},
  year={2024}
}

@inproceedings{sclar2024quantifying,
  title={Quantifying Language Models' Sensitivity to Spurious Features in Prompt Design or: How I learned to start worrying about prompt formatting},
  author={Sclar, Melanie and Choi, Yejin and Tsvetkov, Yulia and Suhr, Alane},
  booktitle={International Conference on Learning Representations},
  volume={2024},
  pages={25055--25083},
  year={2024}
}

@article{cao2024worst,
  title={On the worst prompt performance of large language models},
  author={Cao, Bowen and Cai, Deng and Zhang, Zhisong and Zou, Yuexian and Lam, Wai},
  journal={Advances in Neural Information Processing Systems},
  volume={37},
  pages={69022--69042},
  year={2024}
}

@inproceedings{salinas2024butterfly,
  title={The butterfly effect of altering prompts: How small changes and jailbreaks affect large language model performance},
  author={Salinas, Abel and Morstatter, Fred},
  booktitle={Findings of the Association for Computational Linguistics: ACL 2024},
  pages={4629--4651},
  year={2024}
}

@article{bolton2024rambla,
  title={RAmBLA: A framework for evaluating the reliability of LLMs as assistants in the biomedical domain},
  author={Bolton, William James and Poyiadzi, Rafael and Morrell, Edward R and Bueno, Gabriela van Bergen Gonzalez and Goetz, Lea},
  journal={arXiv preprint arXiv:2403.14578},
  year={2024}
}

@article{guan2025order,
  title={The Order Effect: Investigating Prompt Sensitivity in Closed-Source LLMs},
  author={Guan, Bryan and Roosta, Tanya and Passban, Peyman and Rezagholizadeh, Mehdi},
  journal={arXiv preprint arXiv:2502.04134},
  year={2025}
}

@article{han2024medical,
  title={Medical large language models are susceptible to targeted misinformation attacks},
  author={Han, Tianyu and Nebelung, Sven and Khader, Firas and Wang, Tianci and M{\"u}ller-Franzes, Gustav and Kuhl, Christiane and F{\"o}rsch, Sebastian and Kleesiek, Jens and Haarburger, Christoph and Bressem, Keno K and others},
  journal={NPJ digital medicine},
  volume={7},
  number={1},
  pages={288},
  year={2024},
  publisher={Nature Publishing Group UK London}
}

@inproceedings{yan2025llm,
  title={LLM Sensitivity Evaluation Framework for Clinical Diagnosis},
  author={Yan, Chenwei and Fu, Xiangling and Xiong, Yuxuan and Wang, Tianyi and Hui, Siu Cheung and Wu, Ji and Liu, Xien},
  booktitle={Proceedings of the 31st International Conference on Computational Linguistics},
  pages={3083--3094},
  year={2025}
}

@article{ness2024medfuzz,
  title={Medfuzz: Exploring the robustness of large language models in medical question answering},
  author={Ness, Robert Osazuwa and Matton, Katie and Helm, Hayden and Zhang, Sheng and Bajwa, Junaid and Priebe, Carey E and Horvitz, Eric},
  journal={arXiv preprint arXiv:2406.06573},
  year={2024}
}

@article{moradi2022improving,
  title={Improving the robustness and accuracy of biomedical language models through adversarial training},
  author={Moradi, Milad and Samwald, Matthias},
  journal={Journal of Biomedical Informatics},
  volume={132},
  pages={104114},
  year={2022},
  publisher={Elsevier}
}

@inproceedings{ceballos2024open,
  title={Open (clinical) llms are sensitive to instruction phrasings},
  author={Ceballos-Arroyo, Alberto Mario and Munnangi, Monica and Sun, Jiuding and Zhang, Karen and Mcinerney, Jered and Wallace, Byron C and Amir, Silvio},
  booktitle={Proceedings of the 23rd Workshop on Biomedical Natural Language Processing},
  pages={50--71},
  year={2024}
}

@article{shayegani2023survey,
  title={Survey of vulnerabilities in large language models revealed by adversarial attacks},
  author={Shayegani, Erfan and Mamun, Md Abdullah Al and Fu, Yu and Zaree, Pedram and Dong, Yue and Abu-Ghazaleh, Nael},
  journal={arXiv preprint arXiv:2310.10844},
  year={2023}
}

\end{document}